# Ranking suspected answers to natural language questions using predictive annotation

**Dragomir R. Radev**[*]
School of Information
University of Michigan
Ann Arbor, MI 48103
radev@umich.edu

**John Prager**
TJ Watson Research Center
IBM Research Division
Hawthorne, NY 10532
jprager@us.ibm.com

**Valerie Samn**[*]
Teachers College
Columbia University
New York, NY 10027
vs115@columbia.edu

## Abstract

In this paper, we describe a system to rank suspected answers to natural language questions. We process both corpus and query using a new technique, predictive annotation, which augments phrases in texts with labels anticipating their being targets of certain kinds of questions. Given a natural language question, our IR system returns a set of matching passages, which we then rank using a linear function of seven predictor variables. We provide an evaluation of the techniques based on results from the TREC Q&A evaluation in which our system participated.

## 1 Introduction

Question Answering is a task that calls for a combination of techniques from Information Retrieval and Natural Language Processing. The former has the advantage of years of development of efficient techniques for indexing and searching large collections of data, but lacks of any meaningful treatment of the semantics of the query or the texts indexed. NLP tackles the semantics, but tends to be computationally expensive.

We have attempted to carve out a middle ground, whereby we use a modified IR system augmented by shallow NL parsing. Our approach was motivated by the following problem with traditional IR systems. Suppose the user asks "Where did <some event> happen?". If the system does no pre-processing of the query, then "where" will be included in the bag of words submitted to the search engine, but this will not be helpful since the target text will be unlikely to contain the word "where". If the word is stripped out as a stop-word, then the search engine will have no idea that a location is sought. Our approach, called *predictive annotation*, is to augment the query with semantic category markers (which we call *QA-Tokens*), in this case with the PLACE$ token, and also to label with QA-Tokens all occurrences in text that are recognized entities, (for example, places). Then traditional bag-of-words matching proceeds successfully, and will return matching passages. The answer-selection process then looks for and ranks in these passages occurrences of phrases containing the particular QA-Token(s) from the augmented query. This classification of questions is conceptually similar to the query expansion in (Voorhees, 1994) but is expected to achieve much better performance since potentially matching phrases in text are classified in a similar and synergistic way.

Our system participated in the official TREC Q&A evaluation. For 200 questions in the evaluation set, we were asked to provide a list of 50-byte and 250-byte extracts from a 2-GB corpus. The results are shown in Section 7.

Some techniques used by other participants in the TREC evaluation are paragraph indexing, followed by abductive inference (Harabagiu and Maiorano, 1999) and knowledge-representation combined with information retrieval (Breck et al., 1999). Some earlier systems related to our work are FaqFinder (Kulyukin et al., 1998), MURAX (Kupiec, 1993), which uses an encyclopedia as a knowledge base from which to extract answers, and PROFILE (Radev and McKeown, 1997) which identifies named entities and noun phrases that describe them in text.

## 2 System description

Our system (Figure 1) consists of two pieces: an IR component (GuruQA) that which returns matching texts, and an answer selection compo-

---

nent (AnSel/Werlect) that extracts and ranks potential answers from these texts.

This paper focuses on the process of *ranking potential answers* selected by the IR engine, which is itself described in (Prager et al., 1999).

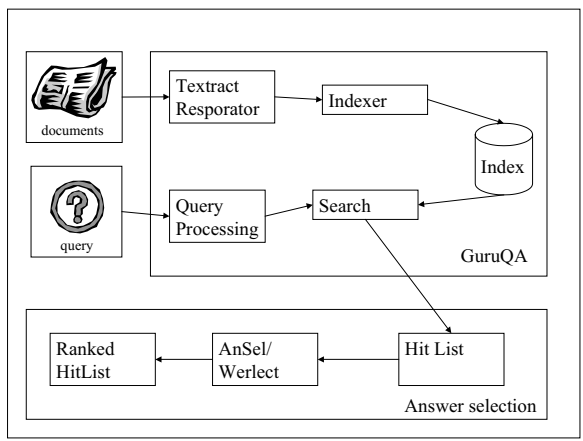

Figure 1: System Architecture.

## 2.1 The Information Retrieval component

In the context of fact-seeking questions, we made the following observations:

- In documents that contain the answers, the query terms tend to occur in close proximity to each other.

- The answers to fact-seeking questions are usually phrases: "President Clinton", "in the Rocky Mountains", and "today").

- These phrases can be categorized by a set of a dozen or so labels (Figure 2) corresponding to question types.

- The phrases can be identified in text by pattern matching techniques (without full NLP).

As a result, we defined a set of about 20 categories, each labeled with its own QA-Token, and built an IR system which deviates from the traditional model in three important aspects.

- We process the query against a set of approximately 200 question templates which, may replace some of the query words with a set of QA-Tokens, called a **SYN-class**. Thus "Where" gets mapped to "PLACE$", but "How long " goes to "@SYN(LENGTH$, DURATION$)". Some templates do not cause complete replacement of the matched string. For example, the pattern "What is the population" gets replaced by "NUMBER$ population".

- Before indexing the text, we process it with Textract (Byrd and Ravin, 1998; Wacholder et al., 1997), which performs lemmatization, and discovers proper names and technical terms. We added a new module (Resporator) which annotates text segments with QA-Tokens using pattern matching. Thus the text "for 5 centuries" matches the DURATION$ pattern "for :CARDINAL _timeperiod", where :CARDINAL is the label for cardinal numbers, and _timeperiod marks a time expression.

- GuruQA scores text passages instead of documents. We use a simple document- and collection-independent weighting scheme: QA-Tokens get a weight of 400, proper nouns get 200 and any other word - 100 (stop words are removed in query processing after the pattern template matching operation). The density of matching query tokens within a passage is contributes a score of 1 to 99 (the highest scores occur when all matched terms are consecutive).

Predictive Annotation works best for *Where*, *When*, *What*, *Which* and *How+adjective* questions than for *How+verb* and *Why* questions, since the latter are typically not answered by phrases. However, we observed that "by" + the present participle would usually indicate the description of a procedure, so we instantiate a METHOD$ QA-Token for such occurrences. We have no such QA-Token for Why questions, but we do replace the word "why" with "@SYN(result, cause, because)", since the occurrence of any of these words usually betokens an explanation.

## 3 Answer selection

So far, we have described how we retrieve relevant passages that may contain the answer to a query. The output of GuruQA is a list of 10 short passages containing altogether a large number (often more than 30 or 40) of potential answers in the form of phrases annotated with QA-Tokens.

| QA-Token | Question type | Example |
|---|---|---|
| PLACE$ | Where | In the Rocky Mountains |
| COUNTRY$ | Where/What country | United Kingdom |
| STATE$ | Where/What state | Massachusetts |
| PERSON$ | Who | Albert Einstein |
| ROLE$ | Who | Doctor |
| NAME$ | Who/What/Which | The Shakespeare Festival |
| ORG$ | Who/What | The US Post Office |
| DURATION$ | How long | For 5 centuries |
| AGE$ | How old | 30 years old |
| YEAR$ | When/What year | 1999 |
| TIME$ | When | In the afternoon |
| DATE$ | When/What date | July 4th, 1776 |
| VOLUME$ | How big | 3 gallons |
| AREA$ | How big | 4 square inches |
| LENGTH$ | How big/long/high | 3 miles |
| WEIGHT$ | How big/heavy | 25 tons |
| NUMBER$ | How many | 1,234.5 |
| METHOD$ | How | By rubbing |
| RATE$ | How much | 50 per cent |
| MONEY$ | How much | 4 million dollars |

Figure 2: Sample QA-Tokens.

### 3.1 Answer ranking

We now describe two algorithms, AnSel and Werlect, which rank the spans returned by GuruQA. AnSel and Werlect[1] use different approaches, which we describe, evaluate and compare and contrast. The output of either system consists of five text extracts per question that contain the likeliest answers to the questions.

### 3.2 Sample Input to AnSel/Werlect

The role of answer selection is to decide which among the spans extracted by GuruQA are most likely to contain the precise answer to the questions. Figure 3 contains an example of the data structure passed from GuruQA to our answer selection module.

The input consists of four items:

- a query (marked with <QUERY> tokens in the example),
- a list of 10 passages (one of which is shown above),
- a list of annotated text spans within the passages, annotated with QA-Tokens, and

- the SYN-class corresponding to the type of question (e.g., "PERSON$ NAME$").

The text in Figure 3 contains five spans (potential answers), of which three ("Biography of Margaret Thatcher", "Hugo Young", and "Margaret Thatcher") are of types included in the SYN-class for the question (PERSON NAME). The full output of GuruQA for this question includes a total of 14 potential spans (5 PERSONs and 9 NAMEs).

### 3.3 Sample Output of AnSel/Werlect

The answer selection module has two outputs: internal (phrase) and external (text passage).

**Internal output:** The internal output is a ranked list of spans as shown in Table 1. It represents a ranked list of the spans (potential answers) sent by GuruQA.

**External output:** The external output is a ranked list of 50-byte and 250-byte extracts. These extracts are selected in a way to cover the highest-ranked spans in the list of potential answers. Examples are given later in the paper.

The external output was required for the TREC evaluation while system's internal output can be used in a variety of applications, e.g., to highlight the actual span that we believe is the answer to the question within the context of the passage in which it appears.

---
[1]from ANswer SELect and ansWER seLECT, respectively

```
<p><NUMBER>1</NUMBER></p>

<p><QUERY>Who is the author of the book, "The Iron Lady: A Biography of Margaret Thatcher"?
</QUERY></p>

<p><PROCESSED_QUERY>@excwin(*dynamic* @weight(200 *Iron_Lady) @weight(200
Biography_of_Margaret_Thatcher) @weight(200 Margaret) @weight(100 author)
@weight(100 book) @weight(100 iron) @weight(100 lady) @weight(100 :) @weight(100 biography)
@weight(100 thatcher) @weight(400 @syn(PERSON$ NAME$)))</PROCESSED_QUERY></p>

<p><DOC>LA090290-0118</DOC></p> <p><SCORE>1020.8114</SCORE></p>

<TEXT><p>THE IRON LADY; A <span class="NAME">Biography of Margaret Thatcher </span>
by <span class="PERSON">Hugo Young</span> (<span class="ORG">Farrar , Straus
& Giroux</span>) The central riddle revealed here is why, as a woman <span class="PLACEDEF">in a
man</span>'s world, <span class="PERSON">Margaret Thatcher</span> evinces such an exclusionary
attitude toward women.</p></TEXT>
```

Figure 3: Input sent from GuruQA to AnSel/Werlect.

| Score | Span |
|---:|---|
| 5.06 | Hugo Young |
| -8.14 | Biography of Margaret Thatcher |
| -13.60 | David Williams |
| -18.00 | Williams |
| -19.38 | Sir Ronald Millar |
| -26.06 | Santiago |
| -31.75 | Oxford |
| -32.38 | Maggie |
| -36.78 | Seriously Rich |
| -42.68 | FT |
| -198.34 | Margaret Thatcher |
| -217.80 | Thatcher |
| -234.55 | Iron Lady |

Table 1: Ranked potential answers to Quest. 1.

## 4 Analysis of corpus and question sets

In this section we describe the corpora used for training and evaluation as well as the questions contained in the training and evaluation question sets.

### 4.1 Corpus analysis

For both training and evaluation, we used the TREC corpus, consisting of approximately 2 GB of articles from four news agencies.

### 4.2 Training set TR38

To train our system, we used 38 questions (see Figure 4) for which the answers were provided by NIST.

### 4.3 Test set T200

The majority of the 200 questions (see Figure 5) in the evaluation set (T200) were not substantially different from these in TR38, although the introduction of "why" and "how" questions as well as the wording of questions in the format "Name X" made the task slightly harder.

| Question/Answer (TR38) |
|---|
| Q: Who was Johnny Mathis' high school track coach? |
| A: Lou Vasquez |
| |
| Q: What year was the Magna Carta signed? |
| A: 1215 |
| |
| Q: What two companies produce bovine somatotropin? |
| A: Monsanto and Eli Lilly |

Figure 4: Sample questions from TR38.

| Question/Answer (T200) |
|---|
| Q: Why did David Koresh ask the FBI for a word processor? |
| A: to record his revelations. |
| |
| Q: How tall is the Matterhorn? |
| A: 14,776 feet 9 inches |
| |
| Q: How tall is the replica of the Matterhorn at Disneyland? |
| A: 147-foot |

Figure 5: Sample questions from T200.

Some examples of problematic questions are shown in Figure 6.

> Q: Why did David Koresh ask the FBI for a word processor?
> Q: Name the first private citizen to fly in space.
> Q: What is considered the costliest disaster the insurance industry has ever faced?
> Q: What did John Hinckley do to impress Jodie Foster?
> Q: How did Socrates die?

Figure 6: Sample harder questions from T200.

## 5 AnSel

AnSel uses an optimization algorithm with 7 predictive variables to describe how likely a given span is to be the correct answer to a question. The variables are illustrated with examples related to the sample question number 10001 from TR38 "Who was Johnny Mathis' high school track coach?". The potential answers (extracted by GuruQA) are shown in Table 2.

### 5.1 Feature selection

The seven span features described below were found to correlate with the correct answers.

**Number:** position of the span among all spans returned from the hit-list.

**Rspanno:** position of the span among all spans returned within the current passage.

**Count:** number of spans of any span class retrieved within the current passage.

**Notinq:** the number of words in the span that do not appear in the query.

**Type:** the position of the span type in the list of potential span types. Example: Type ("Lou Vasquez") = 1, because the span type of "Lou Vasquez", namely "PERSON" appears first in the SYN-class "PERSON ORG NAME ROLE".

**Avgdst:** the average distance in words between the beginning of the span and query words that also appear in the passage. Example: given the passage "Tim O'Donohue, Woodbridge High School's varsity baseball coach, resigned Monday and will be replaced by assistant Johnny Ceballos, Athletic Director Dave Cowen said." and the span "Tim O'Donohue", the value of avgdst is equal to 8.

**Sscore:** passage relevance as computed by GuruQA.

**Number:** the position of the span among all retrieved spans.

### 5.2 AnSel algorithm

The TOTAL score for a given potential answer is computed as a linear combination of the features described in the previous subsection:

$$TOTAL = \sum_i w_i * f_i$$

The algorithm that the training component of AnSel uses to learn the weights used in the formula is shown in Figure 7.

```
For each <question,span> tuple in training
set:
1. Compute features for each span
2. Compute TOTAL score for each span
   using current set of weights
Repeat
   3. Compute performance on training
      set
   4. Adjust weights wi through
      logistic regression
Until performance > threshold
```

Figure 7: Training algorithm used by AnSel.

Training discovered the following weights: $w_{number} = -0.3; w_{rspanno} = -0.5; w_{count} = 3.0; w_{notinq} = 2.0; w_{types} = 15.0; w_{avgdst} = -1.0; w_{sscore} = 1.5$

At runtime, the weights are used to rank potential answers. Each span is assigned a TOTAL score and the top 5 distinct extracts of 50 (or 250) bytes centered around the span are output. The 50-byte extracts for question 10001 are shown in Figure 8. For lack of space, we are omitting the 250-byte extracts.

## 6 Werlect

The Werlect algorithm used many of the same features of phrases used by AnSel, but employed a different ranking scheme.

### 6.1 Approach

Unlike AnSel, Werlect is based on a two-step, rule-based process approximating a function with interaction between variables. In the first stage of this algorithm, we assign a rank to every relevant phrase within each sentence according to how likely it is to be the target answer. Next, we generate and rank each N-byte fragment based on the sentence score given by GuruQA, measures of the fragment's relevance, and the ranks of its component phrases. Unlike AnSel, Werlect was optimized through manual trial-and-error using the TR38 questions.

| Span | Type | Number | Rspanno | Count | Notinq | Type | Avgdst | Sscore | TOTAL |
|---|---|---|---|---|---|---|---|---|---|
| Ollie Matson | PERSON | 3 | 3 | 6 | 2 | 1 | 12 | 0.02507 | -7.53 |
| Lou Vasquez | PERSON | 1 | 1 | 6 | 2 | 1 | 16 | 0.02507 | -9.93 |
| Tim O'Donohue | PERSON | 17 | 1 | 4 | 2 | 1 | 8 | 0.02257 | -12.57 |
| Athletic Director Dave Cowen | PERSON | 23 | 6 | 4 | 4 | 1 | 11 | 0.02257 | -15.87 |
| Johnny Ceballos | PERSON | 22 | 5 | 4 | 1 | 1 | 9 | 0.02257 | -19.07 |
| Civic Center Director Martin Durham | PERSON | 13 | 1 | 2 | 5 | 1 | 16 | 0.02505 | -19.36 |
| Johnny Hodges | PERSON | 25 | 2 | 4 | 1 | 1 | 15 | 0.02256 | -25.22 |
| Derric Evans | PERSON | 33 | 4 | 4 | 2 | 1 | 14 | 0.02256 | -25.37 |
| NEWSWIRE Johnny Majors | PERSON | 30 | 1 | 4 | 2 | 1 | 17 | 0.02256 | -25.47 |
| Woodbridge High School | ORG | 18 | 2 | 4 | 1 | 2 | 6 | 0.02257 | -28.37 |
| Evan | PERSON | 37 | 6 | 4 | 1 | 1 | 14 | 0.02256 | -29.57 |
| Gary Edwards | PERSON | 38 | 7 | 4 | 2 | 1 | 17 | 0.02256 | -30.87 |
| O.J. Simpson | NAME | 2 | 2 | 6 | 2 | 3 | 12 | 0.02507 | -37.40 |
| South Lake Tahoe | NAME | 7 | 5 | 6 | 3 | 3 | 14 | 0.02507 | -40.06 |
| Washington High | NAME | 10 | 6 | 6 | 1 | 3 | 18 | 0.02507 | -49.80 |
| Morgan | NAME | 26 | 3 | 4 | 1 | 3 | 12 | 0.02256 | -52.52 |
| Tennesseefootball | NAME | 31 | 2 | 4 | 1 | 3 | 15 | 0.02256 | -56.27 |
| Ellington | NAME | 24 | 1 | 4 | 1 | 3 | 20 | 0.02256 | -59.42 |
| assistant | ROLE | 21 | 4 | 4 | 1 | 4 | 8 | 0.02257 | -62.77 |
| the Volunteers | ROLE | 34 | 5 | 4 | 2 | 4 | 14 | 0.02256 | -71.17 |
| Johnny Mathis | PERSON | 4 | 4 | 6 | -100 | 1 | 11 | 0.02507 | -211.33 |
| Mathis | NAME | 14 | 2 | 2 | -100 | 3 | 10 | 0.02505 | -254.16 |
| coach | ROLE | 19 | 3 | 4 | -100 | 4 | 4 | 0.02257 | -259.67 |

Table 2: Feature set and span rankings for training question 10001.

| Document ID | Score | Extract |
|---|---|---|
| LA053189-0069 | 892.5 | of O.J. Simpson , Ollie Matson and Johnny Mathis |
| LA053189-0069 | 890.1 | Lou Vasquez , track coach of O.J. Simpson , Ollie |
| LA060889-0181 | 887.4 | Tim O'Donohue , Woodbridge High School 's varsity |
| LA060889-0181 | 884.1 | nny Ceballos , Athletic Director Dave Cowen said. |
| LA060889-0181 | 880.9 | aced by assistant Johnny Ceballos , Athletic Direc |

Figure 8: Fifty-byte extracts.

## 6.2 Step One: Feature Selection

The features considered in Werlect that were also used by AnSel, were Type, Avgdst and Sscore. Two additional features were also taken into account:

**NotinqW:** a modified version of Notinq. As in AnSel, spans that are contained in the query are given a rank of 0. However, partial matches are weighted favorably in some cases. For example, if the question asks, "Who was Lincoln's Secretary of State?" a noun phrase that contains "Secretary of State" is more likely to be the answer than one that does not. In this example, the phrase, "Secretary of State William Seward" is the most likely candidate. This criterion also seems to play a role in the event that Resporator fails to identify relevant phrase types. For example, in the training question, "What shape is a porpoise's tooth?" the phrase "spade-shaped" is correctly selected from among all nouns and adjectives of the sentences returned by Guru-QA.

**Frequency:** how often the span occurs across different passages. For example, the test question, "How many lives were lost in the Pan Am crash in Lockerbie, Scotland?" resulted in four potential answers in the first two sentences returned by Guru-QA. Table 3 shows the frequencies of each term, and their eventual influence on the span rank. The repeated occurrence of "270", helps promote it to first place.

## 6.3 Step two: ranking the sentence spans

After each relevant span is assigned a rank, we rank all possible text segments of 50 (or 250) bytes from the hit list based on the sum of the phrase ranks plus additional points for other words in the segment that match the query.

The algorithm used by Werlect is shown in Figure 9.

| Initial Sentence Rank | Phrase | Frequency | Span Rank |
|---|---|---|---|
| 1 | Two | 5 | 2 |
| 1 | 365 million | 1 | 3 |
| 1 | 11 | 1 | 4 |
| 2 | 270 | 7 | 1 (ranked highest) |

Table 3: Influence of frequency on span rank.

```
1. Let candidate_set = all potential
   answers, ranked and sorted.
2. For each hit-list passage, extract
   all spans of 50 (or 250) bytes, on
   word boundaries.
3. Rank and sort all segments based
   on phrase ranks, matching terms,
   and sentence ranks.
4. For each candidate in sorted
   candidate_set
     - Let highest_ranked_span
       = highest-ranked span
       containing candidate
     - Let answer_set[i++] =
       highest_ranked_span
     - Remove every candidate from
       candidate_set that is found in
       highest_ranked_span
     - Exit if i > 5
5. Output answer_set
```

Figure 9: Algorithm used by Werlect.

## 7 Evaluation

In this section, we describe the performance of our system using results from our four official runs.

### 7.1 Evaluation scheme

For each question, the performance is computed as the reciprocal value of the rank (RAR) of the highest-ranked correct answer given by the system. For example, if the system has given the correct answer in three positions: second, third, and fifth, RAR for that question is $\frac{1}{2}$.

The Mean Reciprocal Answer Rank (MRAR) is used to compute the overall performance of systems participating in the TREC evaluation:

$$RAR = \frac{1}{rank_i}, MRAR = \frac{1}{n}(\sum_i^n \frac{1}{rank_i})$$

### 7.2 Performance on the official evaluation data

Overall, Ansel (runs A50 and A25) performed marginally better than Werlect. However, we noted that on the 14 questions we were unable to classify with a QA-Token, Werlect (runs W50 and W250) achieved an MRAR of 3.5 to Ansel's 2.0.

The cumulative RAR of A50 on T200 (Table 4) is 63.22 (i.e., we got 49 questions among the 198 right from our first try and 39 others within the first five answers).

The performance of A250 on T200 is shown in Table 5. We were able to answer 71 questions with our first answer and 38 others within our first five answers (cumulative RAR = 85.17).

To better characterize the performance of our system, we split the 198 questions into 20 groups of 10 questions. Our performance on groups of questions ranged from 0.87 to 5.50 MRAR for A50 and from 1.98 to 7.5 MRAR for A250 (Table 6).

|  | 50 bytes | 250 bytes |
|---|---|---|
| n | 20 | 20 |
| Avg | 3.19 | 4.30 |
| Min | 0.87 | 1.98 |
| Max | 5.50 | 7.50 |
| Std Dev | 1.17 | 1.27 |

Table 6: Performance on groups of ten questions

Finally, Table 7 shows how our official runs compare to the rest of the 25 official submissions. Our performance using AnSel and 50-byte output was 0.430. The performance of Werlect was 0.395. On 250 bytes, AnSel scored 0.319 and Werlect - 0.280.

## 8 Conclusion

We presented a new technique, **predictive annotation**, for finding answers to natural language questions in text corpora. We showed that a system based on predictive annotation can deliver very good results compared to other competing systems.

We described a set of features that correlate with the plausibility of a given text span being a good answer to a question. We experi-mented with two algorithms for ranking potential answers based on these features. We discovered that a linear combination of these features performs better overall, while a non-linear algorithm performs better on unclassified questions.

|            | First | Second | Third | Fourth | Fifth | TOTAL |
|------------|-------|--------|-------|--------|-------|-------|
| nb of cases | 49   | 15     | 11    | 9      | 4     | 88    |
| Points     | 49.00 | 7.50   | 3.67  | 2.25   | 0.80  | 63.22 |

Table 4: Performance of A50 on T200

|            | First | Second | Third | Fourth | Fifth | TOTAL |
|------------|-------|--------|-------|--------|-------|-------|
| nb of cases | 71   | 16     | 11    | 6      | 5     | 109   |
| Points     | 71.00 | 8.00   | 3.67  | 1.50   | 1.00  | 85.17 |

Table 5: Performance of A250 on T200

| Run  | Median Average | Our Average | Nb Times > Median | Nb Times = Median | Nb Times < Median |
|------|----------------|-------------|-------------------|-------------------|-------------------|
| W50  | 0.12           | 0.280       | 56                | 126               | 16                |
| A50  | 0.12           | 0.319       | 72                | 112               | 14                |
| W250 | 0.29           | 0.395       | 60                | 106               | 32                |
| A250 | 0.29           | 0.430       | 66                | 110               | 22                |

Table 7: Comparison of our system with the other participants

## 9 Acknowledgments

We would like to thank Eric Brown, Anni Coden, and Wlodek Zadrozny from IBM Research for useful comments and collaboration. We would also like to thank the organizers of the TREC Q&A evaluation for initiating such a wonderful research initiative.

## References

Eric Breck, John Burger, David House, Marc Light, and Inderjeet Mani. 1999. Question answering from large document collections. In *Proceedings of AAAI Fall Symposium on Question Answering Systems*, North Falmouth, Massachusetts.

Roy Byrd and Yael Ravin. 1998. Identifying and extracting relations in text. In *Proceedings of NLDB*, Klagenfurt, Austria.

Sanda Harabagiu and Steven J. Maiorano. 1999. Finding answers in large collections of texts : Paragraph indexing + abductive inference. In *Proceedings of AAAI Fall Symposium on Question Answering Systems*, North Falmouth, Massachusetts.

Vladimir Kulyukin, Kristian Hammond, and Robin Burke. 1998. Answering questions for an organization online. In *Proceedings of AAAI*, Madison, Wisconsin.

Julian M. Kupiec. 1993. MURAX: A robust linguistic approach for question answering using an on-line encyclopedia. In *Proceedings, 16th Annual International ACM SIGIR Conference on Research and Development in Information Retrieval*.

John Prager, Dragomir R. Radev, Eric Brown, Anni Coden, and Valerie Samn. 1999. The use of predictive annotation for question answering in TREC8. In *Proceedings of TREC-8*, Gaithersburg, Maryland.

Dragomir R. Radev and Kathleen R. McKeown. 1997. Building a generation knowledge source using internet-accessible newswire. In *Proceedings of the 5th Conference on Applied Natural Language Processing*, pages 221–228, Washington, DC, April.

Ellen Voorhees. 1994. Query expansion using lexical-semantic relations. In *Proceedings of ACM SIGIR*, Dublin, Ireland.

Nina Wacholder, Yael Ravin, and Misook Choi. 1997. Disambiguation of proper names in text. In *Proceedings of the Fifth Applied Natural Language Processing Conference*, Washington, D.C. Association for Computational Linguistics.